\documentclass[journal]{IEEEtran}

\usepackage{cite}
\usepackage{amsmath,amssymb}
\usepackage{graphicx}
\usepackage{url}
\usepackage[colorlinks,urlcolor=blue,linkcolor=blue,citecolor=blue]{hyperref}

\begin{document}

\title{Reinforcement Inference: Leveraging Uncertainty for Self-Correcting Language Model Reasoning}

\author{Xinhai Sun, \IEEEmembership{Member, IEEE}
\thanks{This work was submitted for review on February 9, 2026.}

\thanks{Xinhai Sun is with the Programme of Management Engineering, Politecnico di Milano, Milan, Italy
(e-mail: xinhai.sun@gsom.polimi.it).}
\thanks{Xinhai Sun is also with Synthoid.ai, Shanghai, China
(e-mail: sun.xinhai@synthoid.ai).}}

\maketitle

\begin{abstract}
Modern large language models (LLMs) are often evaluated and deployed under a \emph{one-shot, greedy} inference protocol, especially in professional settings that require deterministic behavior. This regime can systematically under-estimate a fixed model's true capability: many errors arise not from missing knowledge, but from premature commitment under internal ambiguity. We introduce \emph{Reinforcement Inference}, an entropy-aware inference-time control strategy that uses the model's own uncertainty to selectively invoke a second, more deliberate reasoning attempt, enabling stronger performance \emph{without any retraining}.

On 12,032 MMLU-Pro questions across 14 subjects, using DeepSeek-v3.2 with deterministic decoding in a zero-shot setting, Reinforcement Inference improves accuracy from 60.72\% to 84.03\%, while only incurring 61.06\% additional inference calls. A 100\% re-asking ablation reaches 84.35\%, indicating that uncertainty-aware selection captures most of the attainable improvement with substantially less compute. Moreover, a \emph{prompt-only} ablation underperforms the baseline, suggesting that the gains are not explained by generic ``your output had high entropy, think step-by-step'' prompting alone.

Beyond providing a practical inference-time upgrade, our results suggest a broader \emph{entropy-aware} paradigm for measuring and expanding model capability: because modern decoder-based models generate outputs autoregressively, entropy and related confidence measures arise naturally as first-class control signals during generation. The resulting gap between one-pass greedy inference and uncertainty-conditioned deliberation offers a diagnostic lens on an LLM's latent reasoning horizon and motivates future training objectives that explicitly constrain correctness--confidence alignment.
\end{abstract}

\begin{IEEEkeywords}
Large Language Models, Artificial intelligence, Uncertainty Quantification, Self-Correction, Chain-of-Thought, Reinforcement Inference, MMLU-Pro
\end{IEEEkeywords}

\maketitle

\section{Introduction}

\IEEEPARstart{D}{espite} the remarkable performance, LLMs can produce confidently incorrect answers. A well-known challenge is that models often lack calibrated confidence – they may not recognize when their answers are likely wrong \cite{liu2025uncertainty, hendrycks2017baseline}. This can lead to misleading outputs in high-stakes settings (for example, an LLM hallucinating a plausible-sounding but false explanation with full certainty \cite{ji2023survey}). If an AI could know when it is unsure, it might mitigate errors by double-checking or deferring answers. Indeed, prior studies on tasks like multiple-choice QA have noted that an LLM’s softmax probabilities carry information about correctness \cite{hendrycks2017baseline, jiang2020know}. Intuitively, a model that assigns one option nearly 100\% probability is likely confident (and often correct), whereas if it spreads probability roughly uniformly across all options, it is essentially guessing \cite{hendrycks2017baseline}. Recent evaluations confirm that wrong answers tend to have lower maximum softmax probability (MSP) than correct answers in well-trained models \cite{hendrycks2017baseline, guo2017calibration}. In short, uncertainty correlates with mistakes.

How can we use this insight to improve model reasoning? One approach is to have the model abstain or refuse to answer when uncertainty is high \cite{wen2025know}. Refusal might be safer in open-ended applications, but in many scenarios we do want the model to attempt an answer – just with more care if it was initially unsure. Humans do this naturally: if unsure of an answer, a person might pause and double-check their reasoning or try a different approach before responding \cite{madaan2023self}. This inspires the question: Can an LLM similarly “think twice” when it hesitates, to produce a more accurate answer, all at inference time?

\begin{figure}[!t]
\centering
\includegraphics[width=0.5\textwidth]{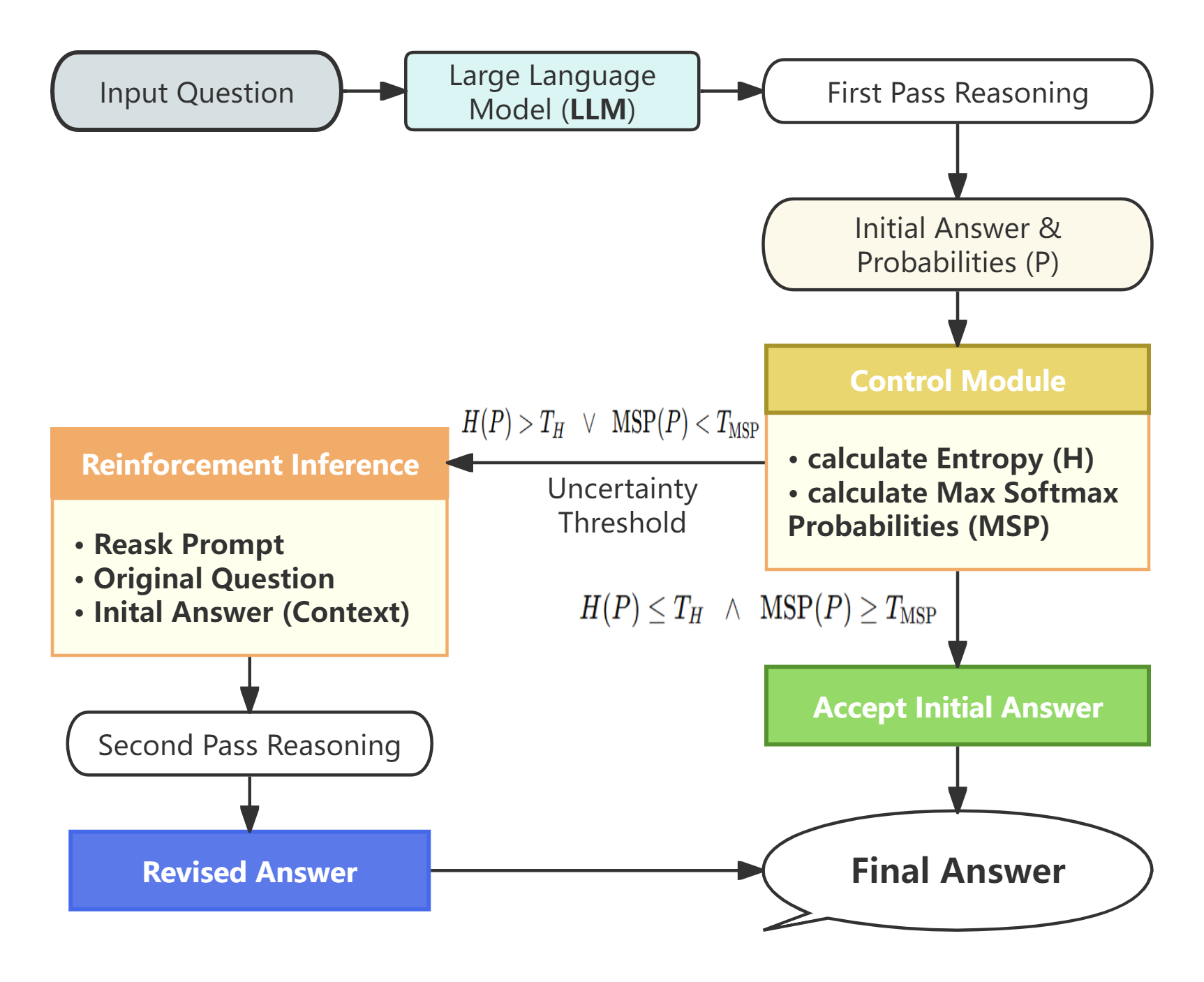}
\caption{System structure of Reinforcement Inference.}
\label{fig:ri_system}
\end{figure}

In this paper, we introduce \emph{Reinforcement Inference}, an inference-time framework that lets an LLM use its own uncertainty to decide when to ``think twice.'' We run a deterministic first pass, monitor entropy and maximum softmax probability (MSP), and trigger a single additional, more deliberate attempt only when the model is uncertain. Crucially, this uses no labels or external feedback and requires \emph{no retraining}.

We argue that \textbf{default one-shot greedy decoding can understate a fixed model's true capability}: in deterministic deployments, many failures reflect premature commitment under ambiguity rather than missing knowledge \cite{madaan2023self}. Reinforcement Inference operationalizes this view by treating uncertainty as a first-class control signal, yielding an adaptive compute policy with predictable behavior on easy cases and extra deliberation only when needed.

More broadly, our results motivate an \textbf{entropy-aware training direction} that explicitly shapes correctness--confidence alignment. The resulting ``capability gap'' between one-pass inference and uncertainty-conditioned deliberation can serve as a practical diagnostic for system design and model optimization.

\section{Contributions}

Our contributions can be summarized as follows:

\begin{itemize}
    \item \textbf{A ``capability-horizon'' view of inference without retraining:} We argue and empirically demonstrate that the default \emph{one-shot, greedy} inference protocol can materially under-estimate what a fixed LLM can do. Reinforcement Inference provides a practical way to unlock this latent reasoning horizon \emph{without any retraining}, which is directly relevant to industrial deployments where model weights cannot be updated frequently.

    \item \textbf{Entropy-aware inference as a systems paradigm:} We introduce an entropy-aware inference-time control strategy that treats uncertainty as a first-class signal for allocating extra compute. This yields a deployment-friendly accuracy--latency trade-off: easy (confident) cases remain single-pass and deterministic, while only uncertain cases trigger an additional deliberate attempt.

    \item \textbf{Evidence that uncertainty is predictive (first-round trial, no leakage):} On 12,032 MMLU-Pro questions, we report a clear separation between correct and incorrect first-pass answers (mean entropy 0.845 vs. 1.567 nats; mean MSP 0.554 vs. 0.249). Importantly, in this \emph{first-round trial} the model never sees any ground-truth labels; correctness is used only for post-hoc evaluation, so there is no ground-truth leakage to the model.

    \item \textbf{Experimental validation + 16-run threshold sweep:} We validate large gains from uncertainty-conditioned re-asking at scale, and we additionally run a 16-configuration sweep of $(\tau_H,\tau_{\mathrm{MSP}})$ to study robustness and identify compute-efficient operating points for future validation. Across the sweep, final accuracy remains stable (about 83.69\%--84.34\%) while re-ask rates vary (52.80\%--76.31\%), demonstrating that the approach is not brittle and enabling practitioners to choose thresholds based on budget.

    \item \textbf{A practical inference-parameter paradigm:} Beyond a single fixed configuration, we frame uncertainty-conditioned re-asking as a tunable \emph{inference policy} with meaningful ``best'' operating regions. In practice, a model can have near-optimal threshold ranges where investing a controlled amount of extra compute yields substantially higher correctness on hard instances, while preserving deterministic behavior and avoiding unnecessary overhead on easy ones.
\end{itemize}

\section{Related Work}

\subsection{Confidence and Uncertainty in LLMs}
Recent work studies how well an LLM’s predictive probabilities reflect correctness and whether models can explicitly reason about their own confidence. Maximum softmax probability (MSP) is a common proxy and tends to be lower for incorrect and out-of-distribution inputs \cite{liu2025uncertainty}. Studies show that softmax distributions can predict correctness in multiple-choice QA and can be used for selective abstention \cite{hendrycks2017baseline, wen2025know}. Related work studies Bayesian/ensemble uncertainty estimation and calibration via fine-tuning \cite{guo2017calibration, gal2016dropout}. We likewise use the model’s output distribution as an uncertainty signal, but rather than training calibration or abstention, we trigger self-correction at inference time.

\subsection{Chain-of-Thought and Self-Consistency}
Chain-of-thought (CoT) prompting can improve performance by eliciting intermediate reasoning \cite{wei2022chain}. Self-consistency extends CoT by sampling multiple reasoning paths and selecting the majority answer \cite{wang2023self}, yielding sizable gains on tasks such as math and commonsense reasoning. However, it is compute-intensive and uses multiple attempts regardless of confidence. In contrast, Reinforcement Inference triggers only a single additional pass and does so selectively when the first pass is uncertain, saving compute on easy queries.

\subsection{Verifier Models and Answer Re-Ranking}
Another line improves correctness via explicit verification. Some methods train separate verifiers to accept/reject model answers, while others prompt the model to verify itself: Chain-of-Verification (CoVe) iteratively generates an answer, asks and answers follow-up checks, and revises the output, reducing hallucinations \cite{dhuliawala2024cove}. Related self-critique approaches similarly analyze and refine the initial output. Compared to these multi-step pipelines, Reinforcement Inference is lightweight: a single uncertainty-triggered second pass without learned verifiers or explicit verification questions.

\subsection{Supervised Retraining for Reasoning}
A common way to improve accuracy or calibration is additional training or fine-tuning on reasoning data or aligning with RLHF. Recent work has also fine-tuned LLMs to output better-calibrated uncertainty or to abstain (``I don't know'') \cite{ouyang2022training}. However, such approaches require labeled data/rewards and retraining, which can be impractical for very large or closed-source models. In contrast, our method requires no training: it is an inference-time plug-in applicable to any model where a confidence signal (e.g., probabilities/logits) can be obtained.

\section{Methodology}

\subsection{Uncertainty Metrics for Model Answers}

We focus on multiple-choice questions, where the model must choose one correct option from a given set (e.g. A, B, C, D, E, F, G, H, I, J). When prompted with a question and answer choices, the model produces a probability distribution \(P\) over the options.

In the main experiments, we compute entropy in nats. For cross-dataset and cross-model evaluation, we report normalized entropy to ensure scale invariance across tasks with different numbers of answer options.

We define two metrics from this distribution:

\begin{itemize}
    \item \textbf{Entropy (\(H\)):} This measures the overall uncertainty or spread of the distribution. Formally, $$H(P) = -\sum_{i=1}^{K} P_i \ln P_i,$$ where \(K\) is the number of options (\(K=10\) in MMLU-Pro). We express entropy in nats; in this case, a uniform distribution has \(H = \ln K\) (e.g., \(H=\ln 10\approx 2.303\) when \(K=10\)), while a peaked distribution (e.g. one option with \(P \approx 1\) and the others near 0) has entropy close to 0. In our analysis, we compute \(H\) using the model’s normalized probabilities over the provided choices. (In the extreme deterministic case, where $p_{\max}=1$, we have $H=-\log p_{\max}$.)
    \item \textbf{Maximum Softmax Probability (\(MSP\)):} This is the model’s confidence in its top choice, defined as $$\mathrm{MSP}(P) = \max_{1\le i\le K} P_i.$$ For example, if the model selects option A with probability 0.7, then \(\mathrm{MSP}=0.7\). A higher \(\mathrm{MSP}\) indicates the model was more confident in the answer it chose, whereas a low \(\mathrm{MSP}\) (e.g. 0.25) indicates it was essentially guessing among several options.
    \item \textbf{Normalized Entropy (\(\bar{H}\)):} Because \(H\) scales with \(K\), we also consider a normalized form
    $$\bar{H}(P) = \frac{H_{\mathrm{nat}}(P)}{\ln K} \in [0,1],$$
    which enables comparing uncertainty across tasks with different numbers of options. This normalization is a monotonic rescaling and does not alter the triggering behavior.
\end{itemize}

These metrics are inversely related (a high \(\mathrm{MSP}\) usually implies low entropy, and vice versa), but we consider both for completeness. Entropy captures the notion of hesitation in the distribution, while \(\mathrm{MSP}\) directly captures confidence in the chosen answer. From a Bayesian decision perspective, if the model’s posterior over options is \(P\), the Bayes-optimal 0--1 decision is \(\hat{y} = \arg\max_i P_i\), and \(1-\mathrm{MSP}(P)\) upper-bounds the posterior risk of choosing \(\hat{y}\) under the model’s own beliefs. Inspired by classical post-hoc calibration methods such as Platt scaling \cite{platt1999probabilistic}, we also define two proper scoring rules \cite{gneiting2007scoring}, which we report as secondary diagnostics in Section~\ref{sec:uncertainty_metrics}:

\begin{itemize}
    \item \textbf{Brier score:} for one-hot label \(y\), $$\mathrm{BS}(P,y) = \sum_{i=1}^{K} (P_i - \mathbb{I}[i=y])^2,$$ which penalizes both overconfidence and underconfidence.
    \item \textbf{Negative log-likelihood (NLL):} $$\mathrm{NLL}(P,y) = -\log P_y,$$ which corresponds to the cross-entropy loss of the true option.
\end{itemize}

We hypothesize that correct answers will on average have lower entropy and higher \(\mathrm{MSP}\) than incorrect ones – indicating the model tends to be more decisive when it is right, and more ambivalent when it is wrong.

To test this hypothesis, we ran DeepSeek-v3.2 on a large set of multiple-choice questions and extracted the model’s answer distribution over the provided options from token log-probabilities (converted to normalized probabilities). We then computed \(H\) and \(\mathrm{MSP}\) for each question and compared these uncertainty metrics between items that were answered correctly vs. incorrectly \emph{post hoc}.

This analysis is inspired by standard calibration-style evaluation practice: rather than claiming any specific distributional assumptions, we report descriptive statistics and visualize the separation between the two groups.

\subsection{Empirical Trigger Criterion}

We use uncertainty to decide whether to invoke a second reasoning attempt. We compute entropy \(H(P)\) using the natural logarithm, yielding values in nats. For ten-way multiple-choice questions (MMLU-Pro), the maximum entropy is \(\ln 10 \approx 2.303\) nats.

\textbf{Main setting (MMLU-Pro, 10-way).} To avoid any threshold selection leakage, our main setting uses fixed, a priori thresholds motivated by the interpretation of entropy/MSP under a Bayesian 0--1 decision framework. For $K=10$ options, the maximum entropy is $\ln K\approx 2.303$ nats.

We adopt an entropy threshold of $\tau_H=0.8$ nats, corresponding to roughly 35\% of the maximum uncertainty,
$$\frac{\tau_H}{\ln 10} \approx 0.347,$$
which selects predictions with a clearly peaked but not fully deterministic posterior. In parallel, we require $\mathrm{MSP}(P)\ge \tau_{\mathrm{MSP}}=0.6$, which upper-bounds the model’s posterior risk by $1-\tau_{\mathrm{MSP}}=0.4$.

Together, these thresholds identify predictions that are both locally confident in the chosen answer and globally low in distributional ambiguity, while avoiding overly conservative or near-deterministic regimes.

We trigger re-asking when either criterion indicates hesitation:
$$\text{trigger} = \mathbb{I}[H(P) > \tau_H\ \lor\ \mathrm{MSP}(P) < \tau_{\mathrm{MSP}}].$$

\textbf{Cross-dataset transfer (MMLU, 4-way).} When transferring the same normalized-entropy threshold to 4-way multiple-choice tasks, we use the equivalent unnormalized entropy threshold
$$\tau_H^{(K=4)} = \bar{H}_\tau \ln 4 \approx 0.78\ \text{nats},$$
while keeping \(\tau_{\mathrm{MSP}}=0.4\) fixed.

Although these thresholds were guided by an initial trial analysis, we emphasize that selecting \(\tau_H,\tau_{\mathrm{MSP}}\) after inspecting performance can introduce optimistic bias. We discuss this ``privileged'' threshold selection issue and its implications in Section~\ref{sec:threshold_sweep} and Limitations.

The Reinforcement Inference procedure can be summarized as Algorithm 1:

\begin{enumerate}
    \item \textbf{First Pass:} Present the question to the model in the standard zero-shot format and obtain its answer (along with the probability distribution \(P\) over answer options).
    \item \textbf{Uncertainty Check:} Compute entropy \(H(P)\) and maximum softmax probability \(\mathrm{MSP}(P)\). If \(H(P) \le \tau_H\) and \(\mathrm{MSP}(P) \ge \tau_{\mathrm{MSP}}\), consider the model confident and accept its first answer as final.
    \item \textbf{Reinforcement Prompt (Second Pass):} Otherwise, perform a second query to the model. Prepend a special instruction indicating that the previous attempt showed uncertainty and encouraging more careful, step-by-step reasoning, then provide the same question again.
    \item \textbf{Final Answer:} If a second pass was triggered, take the model’s second answer as the final answer for the question. Otherwise, use the first-pass answer.
\end{enumerate}

Notably, in step 3 we do not give away any information about whether the first answer was correct, nor do we provide any external hints – we only indicate that the model itself was not confident and should rethink. This ensures we are not “cheating” by using ground truth; the model has to solve the problem on its own, albeit with a nudge to be more meticulous.

\subsection{Second-Pass Prompting Protocol}

A critical component of Reinforcement Inference is the design of the second-pass prompt. The goal is to make the model aware that its prior attempt was uncertain, while avoiding any task-specific hints. In our implementation, the second-pass input consists of three parts:

\begin{enumerate}
    \item \textbf{Original question block:} the same question and options in the same structure as the first pass (e.g., \texttt{Question: ...}, \texttt{Options: ...}).
    \item \textbf{Previous answer:} a header line \texttt{Your previous answer was:} followed by the full first-round model output.
    \item \textbf{Instruction:} a header line \texttt{Instruction:} followed by a fixed prompt indicating uncertainty (high entropy) and asking the model to discard the prior answer and recompute from first principles.
\end{enumerate}

Importantly, the reprompt never tells the model it was wrong and never reveals the correct option. We only include the first-pass output as context and request a fresh derivation under an explicit uncertainty cue, so the second pass re-solves the same question with added pressure to be careful (analogous to asking a student to redo a problem after noting hesitation).

In summary, this is a self-reinforcement loop: uncertainty triggers additional reasoning without external hints, preserving a clean evaluation and attributing gains to better use of the model's existing knowledge rather than information leakage.

\section{Experimental Setup}

\subsection{Model and Experiment Settings}

We conducted experiments with DeepSeek-v3.2, a state-of-the-art LLM released in late 2025. It is an open-source model with approximately 685 billion parameters, trained on a massive multilingual and technical corpus and strong across benchmarks \cite{deepseek2025nature}. We chose it as a strong yet challenging testbed for uncertainty-based Reinforcement Inference. Table~\ref{tab:inference_params} summarizes the inference parameters used across all experiments.

We evaluate in a zero-shot, deterministic setting with greedy decoding (temperature $t=0$) to ensure stable outputs and probabilities. Both passes use the same multiple-choice format (question + options A--J) and prompt the model to output answering process and option letter \cite{brown2020language}; the second pass simply prepends the reinforcement instruction. 

To obtain answer probabilities, we query the API for token-level log-probabilities. Because the answer is a single token (A--J), we directly read out the probability of each option, normalize over the ten choices, and compute the entropy $H$ and $MSP$ as described earlier.

\begin{table}[!t]
\caption{Inference parameters.}
\label{tab:inference_params}
\centering
\footnotesize
\setlength{\tabcolsep}{4pt}
\begin{tabular}{ll}
\hline\hline
Parameter & Value \\
\hline
Model & deepseek-v3.2 \\
Temperature & 0 \\
Max tokens & 4000 \\
Top-$p$ & 1 \\
Frequency penalty & 0 \\
Presence penalty & 0 \\
Logprobs & True \\
Top logprobs & 20 \\
\hline\hline
\end{tabular}
\end{table}

\subsection{Dataset and Questions}

We drew our evaluation questions from MMLU-Pro, which is an enhanced subset of the Massive Multitask Language Understanding benchmark focusing on professional or expert-level exams \cite{wang2024mmlu, hendryckstest2021}. MMLU is a collection of 57 subject areas (history, math, science, law, medicine, etc.) with questions ranging from elementary level to professional certification level. The “Pro” subset specifically includes domains like medical board exams, law school exams, advanced mathematics, and other graduate-level or professional topics. These questions are known to be especially challenging for LLMs – models often perform near chance on some professional subjects and, crucially, do not know when they are wrong \cite{srivastava2023beyond}.

\subsection{Reference Results on MMLU-Pro}
To contextualize the difficulty of MMLU-Pro, Table~\ref{tab:mmlupro_reference} reproduces a subset of the reference results reported in the MMLU-Pro paper (CoT, accuracies in \%) \cite{wang2024mmlu}.

\begin{table}[!t]
\caption{A subset of reference MMLU-Pro results reported by \cite{wang2024mmlu} (accuracies in \%).}
\label{tab:mmlupro_reference}
\centering
\footnotesize
\setlength{\tabcolsep}{3pt}
\begin{tabular*}{\linewidth}{@{\extracolsep{\fill}}lrrrrrrr}
\hline\hline
Model & Overall & Math & Physics & Eng. & Hist. & Law & Psych. \\
\hline
GPT-4o & 72.6 & 76.1 & 74.7 & 55.0 & 70.1 & 51.0 & 79.2 \\
Gemini-1.5-Pro & 69.0 & 72.8 & 70.4 & 48.7 & 65.6 & 50.8 & 77.2 \\
Claude-3-Opus & 68.5 & 69.6 & 69.7 & 48.4 & 61.4 & 53.5 & 76.3 \\
GPT-4-Turbo & 63.7 & 62.8 & 61.0 & 35.9 & 67.7 & 51.2 & 78.3 \\
Llama-3-70B-Instruct & 56.2 & 54.0 & 49.6 & 43.6 & 56.9 & 39.9 & 70.2 \\
\hline\hline
\end{tabular*}
\end{table}

From the MMLU-Pro repository, we evaluated 12,032 multiple-choice questions across 14 professional subjects: biology, business, chemistry, computer science, economics, engineering, health, history, law, math, other, philosophy, physics, and psychology \cite{wang2024mmlu}. Each question provided ten answer choices (A through J). All questions were drawn from the official test set of MMLU-Pro; the correct answers were hidden from the model (we only used them to check accuracy after the model produced its answer).

Out of these 12,032 questions, DeepSeek-v3.2 answered 7,306 correctly and 4,726 incorrectly on the first pass. This 60.72\% accuracy is in line with expectations for a strong model on difficult zero-shot questions--leaving ample room for improvement. Using our entropy and max-probability thresholds, 7,347 questions (61.06\%) were flagged as high uncertainty and received a second pass; the remaining 4,685 (38.94\%) were accepted as confident and not re-asked.

\subsection{Procedure}

Our empirical study consists of four experiments that share a common inference pipeline: for each question, we run a deterministic \emph{first pass} to obtain (i) the model’s answer and (ii) token log-probabilities over the answer options, from which we compute the uncertainty metrics (entropy, MSP, and scoring-rule diagnostics). We then optionally run a \emph{second pass} (re-ask) depending on the experimental condition.

\textbf{Experiment I (effectiveness and causality).} We evaluate our main method and two ablations on the same MMLU-Pro test set:
\begin{itemize}
    \item \textbf{Baseline (one-shot):} run only the first pass and use its answer as final.
    \item \textbf{Targeted re-asking (TR, main method):} run the first pass, compute uncertainty, and trigger exactly one second pass when the trigger criterion is satisfied; the second-pass answer becomes the final answer.
    \item \textbf{Uniform re-asking (UR, ablation):} run a second pass for every question (100\%), regardless of uncertainty.
    \item \textbf{Prompt-only re-asking (ablation):} run a single pass using the same ``re-ask'' prompt template as the second pass, but without performing an actual first-pass attempt (and thus without any uncertainty measurement or answer-conditioning). This isolates whether the observed gains could be explained purely by the stronger re-ask prompt wording itself.
\end{itemize}

\textbf{Experiment II (generalization across model architectures and reasoning paradigms).} We extend evaluation to additional state-of-the-art model families (DeepSeek-v3.2-Reasoner, Qwen3-30B-Instruct, and Qwen3-235B-Instruct) to test whether uncertainty-triggered re-asking transfers across (i) distinct reasoning paradigms and (ii) different model architectures and parameter scales, while maintaining inference parameter settings in Experiment I. We keep MMLU-Pro as the testbed for comparability and reuse the same static thresholds as in Experiment I ($\tau_H = 1.3$ nats, $\tau_{\mathrm{MSP}} = 0.4$) to probe robustness across models.

\textbf{Experiment III (robustness / compute trade-off).} We repeat the targeted re-asking policy over a grid of threshold pairs $(\tau_H,\tau_{\mathrm{MSP}})$. For each pair, we record (i) the re-ask rate (fraction of questions triggering a second pass) and (ii) the resulting overall accuracy, producing an empirical accuracy--compute trade-off curve.

\textbf{Experiment IV (cross-dataset transfer).} We transfer the entropy threshold across tasks with different numbers of options by mapping a normalized entropy threshold from 10-way MMLU-Pro to the corresponding unnormalized threshold for 4-way MMLU. We then run the same targeted re-asking procedure on standard MMLU and report transfer performance.

Across all experiments, we evaluate final correctness using ground-truth labels \emph{only after} the model has produced its outputs; no ground-truth information is ever shown to the model during inference.

\subsection{Statistical Tests}

In addition to uncertainty metrics from the predictive distribution, we assess confidence using proper scoring rules, which evaluate probabilistic predictions against the ground-truth label and are minimized in expectation by the true distribution.

We use the Brier score and negative log-likelihood (NLL). The Brier score is the squared error between the predicted probability vector and the one-hot label, while NLL (cross-entropy at the true label) strongly penalizes assigning low probability to the correct option. These complement entropy/MSP by measuring how well confidence aligns with empirical correctness.

To quantify the benefit of Reinforcement Inference, we focus on re-asked (high-uncertainty) questions and apply McNemar’s test to paired first- vs.\ second-pass binary outcomes. Let \(b\) be the number of wrong\(\to\)correct changes and \(c\) the number of correct\(\to\)wrong changes. The McNemar test statistic is
$$\chi^2 = \frac{(|b-c|-1)^2}{b+c},$$
which is approximately \(\chi^2_1\)-distributed under the null (continuity correction shown). We report the corresponding p-value for the high-uncertainty group.

Additionally, we report the raw improvement in accuracy (in percentage points) and compute an effect size for the change. Owing to the large evaluation set of 12,032 questions spanning diverse subjects, aggregate statistics naturally average out idiosyncratic fluctuations at the individual-item level, reducing the likelihood that the observed effects are driven by accidental alignment between a particular parameter setting and a small subset of questions. This scale enables robust paired tests and reliable subject-wise analysis.

\section{Results}

\textbf{Experiments I} evaluate effectiveness and causality (baseline vs. targeted re-asking vs. uniform re-asking). \textbf{Experiments II} evaluate generalization across model architectures and reasoning paradigms under fixed thresholds. \textbf{Experiments III} analyze robustness and the accuracy--compute trade-off via a 16-run threshold sweep. \textbf{Experiments IV} evaluate zero-shot transfer to standard MMLU by mapping the 10-way entropy threshold to a 4-way equivalent via normalized entropy.

\subsection{Model Uncertainty vs. Answer Correctness (Baseline)}
\label{sec:uncertainty_metrics}

\begin{figure}[!t]
\centering
\includegraphics[width=\linewidth]{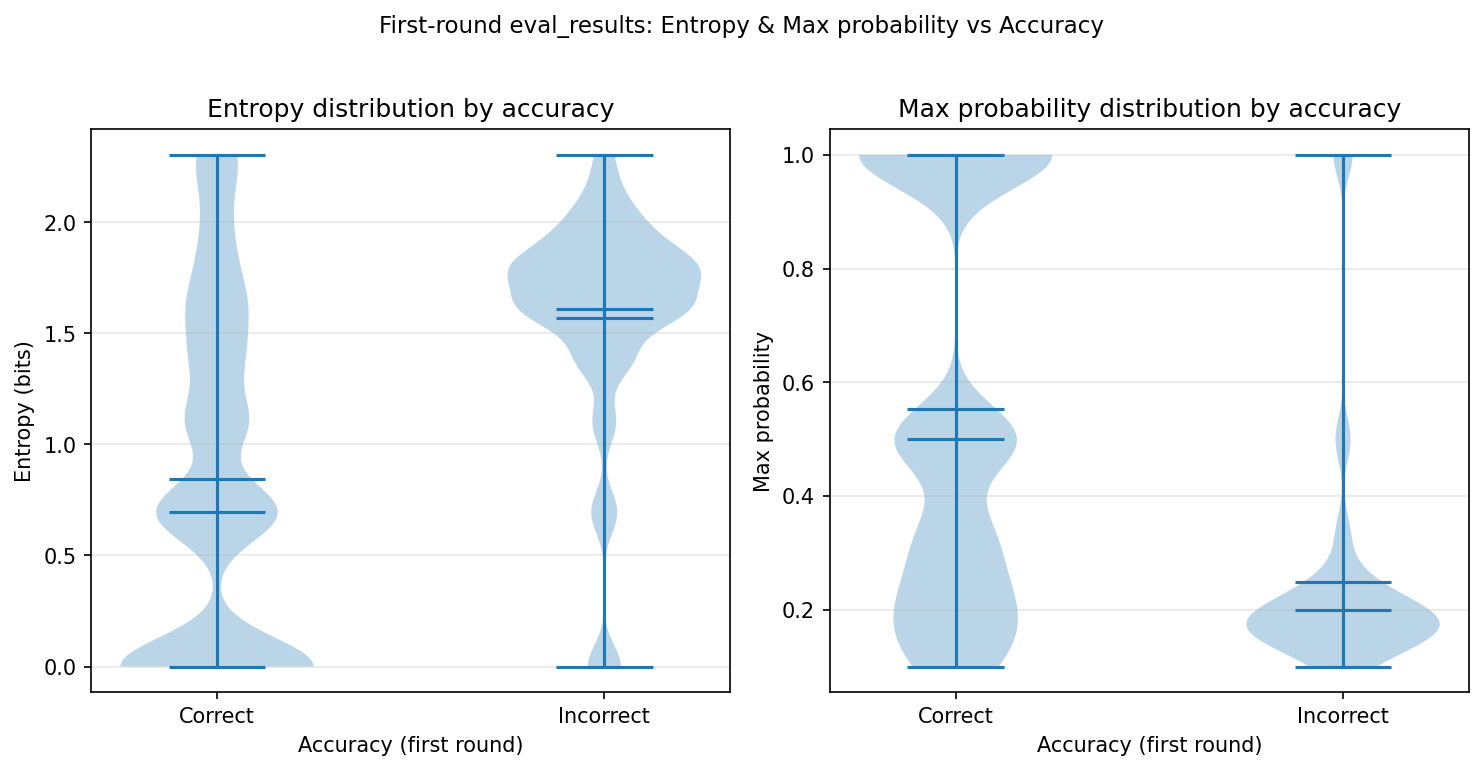}
\caption{First-round distribution of entropy (nats) and MSP on the 12,032-question MMLU-Pro, illustrating the separation between correct and incorrect answers.}
\label{fig:uncertainty}
\end{figure}

We first analyze the \emph{baseline} one-shot pass and ask whether the model’s own uncertainty correlates with correctness. We find a clear separation: DeepSeek-v3.2 is typically more uncertain on questions it answers incorrectly. Figure~\ref{fig:uncertainty} summarizes this baseline relationship between uncertainty (entropy and maximum probability) and correctness \cite{deepseek2025nature}.

From our calculations on 12,032 questions (baseline pass), when the model answered incorrectly, its output distribution was typically much more spread out. The average entropy for correct answers was $0.844$ nats, compared to $1.567$ nats for incorrect answers. Similarly, the maximum probability assigned to the chosen answer was on average much higher for correct answers ($0.554$) than for incorrect answers ($0.249$). In short, baseline uncertainty is strongly predictive of mistakes.

\begin{table}[!t]
\caption{Descriptive statistics of uncertainty metrics (MMLU-Pro, 12,032 questions). Values are mean $\pm$ standard deviation.}
\label{tab:first_round_stats}
\centering
\footnotesize
\setlength{\tabcolsep}{4pt}
\begin{tabular}{llcc}
\hline\hline
Round & Metric & Correct & Incorrect \\
\hline
Baseline & Entropy (nats) & $0.844456\pm0.763501$ & $1.566893\pm0.512095$ \\
TR & Entropy (nats) & $0.940312\pm0.881968$ & $1.261358\pm0.877025$ \\
Baseline & MSP & $0.553804\pm0.349872$ & $0.249250\pm0.206929$ \\
TR & MSP & $0.539224\pm0.365690$ & $0.409469\pm0.339603$ \\
\hline\hline
\end{tabular}
\end{table}

\begin{table}[!t]
\caption{Descriptive statistics of proper scoring rules (lower is better). Values are mean $\pm$ standard deviation.}
\label{tab:first_round_scoring_rules}
\centering
\footnotesize
\setlength{\tabcolsep}{4pt}
\begin{tabular}{llcc}
\hline\hline
Round & Metric & Correct & Incorrect \\
\hline
Baseline & Brier score & $0.446196\pm0.349872$ & $0.965031\pm0.322060$ \\
TR & Brier score & $0.460776\pm0.365690$ & $1.214988\pm0.486595$ \\
Baseline & NLL & $0.844456\pm0.763501$ & $6.563630\pm8.919788$ \\
TR & NLL & $0.940312\pm0.881968$ & $11.663336\pm10.542678$ \\
\hline\hline
\end{tabular}
\end{table}

\begin{table*}[!t]
\caption{Subject-wise re-asking coverage and accuracy on MMLU-Pro. ``1st Acc.'' and ``2nd Acc.'' are the overall accuracies within each subject (computed over all questions in that subject), using the first-pass answer vs. the final TR answer after uncertainty-triggered re-asking. ``Re-asked'' and ``Re-asked (\%)'' report how often TR triggers within each subject; ``Change'' is the absolute improvement in percentage points.}
\label{tab:subject_accuracy}
\centering
\scriptsize
\setlength{\tabcolsep}{4pt}
\renewcommand{\arraystretch}{1.05}
\begin{tabular*}{\textwidth}{@{\extracolsep{\fill}}lrrrrrr}
\hline\hline
Subject & Total & Re-asked & Re-asked (\%) & 1st Acc. (overall, \%) & 2nd Acc. (overall, \%) & Change (pp) \\
\hline
Biology & 717 & 599 & 83.54 & 65.13 & 92.47 & +27.34 \\
Business & 789 & 199 & 25.22 & 80.61 & 87.33 & +6.72 \\
Chemistry & 1132 & 505 & 44.61 & 78.98 & 88.16 & +9.19 \\
Computer Sci. & 410 & 294 & 71.71 & 53.90 & 86.59 & +32.68 \\
Economics & 844 & 715 & 84.72 & 52.25 & 87.44 & +35.19 \\
Engineering & 969 & 570 & 58.82 & 75.44 & 80.08 & +4.64 \\
Health & 818 & 346 & 42.30 & 76.41 & 80.56 & +4.16 \\
History & 381 & 379 & 99.48 & 14.44 & 74.54 & +60.10 \\
Law & 1101 & 1083 & 98.37 & 16.26 & 66.30 & +50.05 \\
Math & 1351 & 738 & 54.63 & 57.29 & 92.75 & +35.46 \\
Other & 924 & 308 & 33.33 & 80.19 & 80.30 & +0.11 \\
Philosophy & 499 & 478 & 95.79 & 43.09 & 81.16 & +38.08 \\
Physics & 1299 & 494 & 38.03 & 77.29 & 88.99 & +11.70 \\
Psychology & 798 & 639 & 80.08 & 40.48 & 83.08 & +42.61 \\
\hline\hline
\end{tabular*}
\end{table*}

In addition to entropy and MSP, proper scoring rules also separate correct from incorrect answers under the baseline pass: correct items have lower Brier score and lower NLL on average.

These baseline results support our key premise at scale: DeepSeek-v3.2’s hesitation signals---higher entropy and lower maximum probability---are strongly associated with mistakes on MMLU-Pro. This behavior is consistent with prior calibration findings on modern neural models \cite{hendrycks2017baseline, guo2017calibration}.

\subsection{Impact of Reinforcement Inference on Accuracy (Targeted Re-asking, TR)}

We now turn to the main evaluation: did the second-pass "reinforcement" prompt actually lead to more correct answers on the flagged uncertain questions? The results confirm this hypothesis significantly.

Out of the 7,347 high-uncertainty questions where we triggered a second pass (TR), the model’s accuracy on that subset rose from 42.62\% on first pass to 80.79\% on second pass--an absolute gain of +38.18 percentage points. In terms of transitions: 2,912 questions improved from wrong to correct (39.64\% of re-asked questions), 107 degraded from correct to wrong (1.46\%), 3,024 stayed correct (41.16\%), and 1,304 stayed wrong (17.75\%). Thus the net change was +2,805 correct answers among the re-asked subset. The 4,685 questions that were not re-asked (those the model was confident about) maintained 89.11\% accuracy on first pass and were left unchanged. Table~\ref{tab:subject_accuracy} further breaks down re-asking coverage and before/after accuracy within each of the 14 subjects. The gains are observed across a wide range of subjects, supporting that TR is a cross-subject inference policy rather than a procedure tuned to fit any single subject.

This improvement is highly significant. By McNemar’s test on the re-asked subset, we observe $b=2912$ (wrong$\to$correct) and $c=107$ (correct$\to$wrong), yielding
$$\chi^2 = \frac{(|b-c|-1)^2}{b+c} = 2604.3114,\qquad p<10^{-6}.$$
In addition to statistical significance, the effect size is large: the re-asked subset improves by +38.18 percentage points (42.62\% $\to$ 80.79\%), with Cohen’s $h=0.8117$.

Across all 12,032 questions, overall accuracy rose from 60.72\% on first pass to 84.03\% after applying Reinforcement Inference, an absolute gain of +23.31 percentage points. The not re-asked subset (38.94\% of questions) contributed its already-high 89.11\% accuracy; the re-asked subset (61.06\%) contributed the large improvement from 42.62\% to 80.79\%. A small number of initially correct answers (107) were flipped to wrong on second pass, but the net benefit was strongly positive.

Table~\ref{tab:first_round_scoring_rules} reports proper scoring rules (Brier score and negative log-likelihood, NLL) before and after uncertainty-triggered re-asking. Although targeted re-asking (TR) improves accuracy, Brier and NLL can increase on the remaining incorrect cases. This is expected: TR mainly fixes moderate-uncertainty mistakes, leaving a smaller set of intrinsically hard errors where the model assigns near-zero probability to the true option, which these losses penalize strongly.

Overall, Reinforcement Inference translates uncertainty awareness into large accuracy gains via effective self-correction, suggesting that many initially missed answers are latent and can be recovered with a second-pass prompt.

\subsection{Ablation Study: Uniform Re-asking (UR)}

To test whether the benefit of Reinforcement Inference comes from targeting uncertain questions or simply from re-asking more questions, we ran a uniform re-asking ablation (UR) in which we re-asked all 12,032 questions (100\%) regardless of entropy or max probability. The results are summarized in Table \ref{tab:ra1_vs_ra2}.

\begin{table}[!t]
\caption{Targeted re-asking (TR) vs. uniform re-asking (UR).}
\label{tab:ra1_vs_ra2}
\centering
\footnotesize
\setlength{\tabcolsep}{4pt}
\begin{tabular}{lll}
\hline\hline
Metric & TR (Targeted) & UR (Uniform) \\
\hline
Questions re-asked & 7,347 (61.06\%) & 12,032 (100\%) \\
Overall accuracy change & +23.31\% & +23.63\% \\
Final overall accuracy & 84.03\% & 84.35\% \\
Computational cost & 1.61$\times$ baseline & 2$\times$ baseline \\
\hline\hline
\end{tabular}
\end{table}

\begin{table}[!t]
\caption{Summary of our inference-time evaluation on 12,032 MMLU-Pro items: mean entropy (nats) and mean maximum softmax probability (MaxProb) across inference rounds.}
\label{tab:our_results}
\centering
\footnotesize
\setlength{\tabcolsep}{4pt}
\begin{tabular}{lrr}
\hline\hline
Round & Entropy mean & MaxProb mean \\
\hline
Baseline & 1.1282 & 0.4342 \\
TR & 0.9916 & 0.5185 \\
UR & 1.1651 & 0.4463 \\
\hline\hline
\end{tabular}
\end{table}

When we re-asked every question (UR), overall accuracy improved by an additional +0.32 percentage points over the targeted approach (TR). In other words, re-asking only the 61.06\% of questions flagged as uncertain achieved 98.6\% of the maximum possible improvement (23.31\% vs 23.63\% gain). Re-asking the remaining 38.94\% of questions (those where the model was already confident) provided negligible benefit. This confirms that entropy-based selection identifies exactly the questions that need re-asking: the model’s internal uncertainty is the key signal, not the volume of re-asking. It is not that the model cannot decide; it is that when the model answers under low confidence (high entropy), re-asking unlocks better reasoning. Addressing uncertainty is therefore central to addressing hallucination when the LLM is confused.

\subsection{Ablation: Prompt-Only Re-asking vs. Uncertainty-Conditioned Re-asking}
\label{sec:prompt_only_ablation}

An alternative explanation for the observed gains is that they arise purely from prompt engineering effects, such as explicitly encouraging step-by-step reasoning or inducing a stronger chain-of-thought (CoT), rather than from the proposed uncertainty-conditioned re-asking mechanism. To disentangle these factors, we conduct an additional ablation that isolates the effect of the re-asking prompt itself.

\textbf{Experimental setup.} We compare the standard zero-shot baseline (DeepSeek-v3.2 evaluated on MMLU-Pro) against a prompt-only re-asking condition. In this ablation, the model is evaluated in a single pass under the same decoding and evaluation settings as the baseline, but with a modified system prompt:

\begin{quote}
The following are multiple choice questions (with answers) about \{\}. Your last output had high entropy, indicating uncertainty. Discard the prior answer and recompute the solution from first principles. Think step by step and then output the answer in the format of ``The answer is (X)'' at the end.
\end{quote}

This condition does not include (i) the model’s first-round prediction, nor (ii) any actual uncertainty signal derived from a prior output, since no initial pass is performed. Thus, this ablation controls for both answer feedback and selective triggering, leaving only the effect of the re-asking instruction and CoT encouragement.

\textbf{Results.} The prompt-only re-asking condition underperforms the zero-shot baseline. That is, merely instructing the model to ``discard the prior answer'' and ``recompute from first principles,'' even with explicit step-by-step reasoning, does not improve---and in fact degrades---performance.

\textbf{Implications.} This indicates that gains do not come from the prompt or generic CoT elicitation alone; they require uncertainty-triggered, selective re-asking conditioned on a concrete first-round attempt. In short, the key is uncertainty-aware, answer-conditioned re-asking rather than urging the model to ``think harder.''

\subsection{Experiment II: Generalization Across Model Architectures and Reasoning Paradigms}
\label{sec:exp2_generalization}

\textbf{Experimental setup.} While our primary results demonstrate the efficacy of Reinforcement Inference on DeepSeek-v3.2, it remains unclear whether uncertainty-triggered self-correction is specific to one model family. We therefore evaluate three additional models with distinct characteristics: DeepSeek-v3.2-Reasoner (reasoning-oriented inference), Qwen3 30B, and Qwen3 235B (different architectures and scales). We keep MMLU-Pro as the testbed, while maintaining inference parameter settings in Experiment I, and reuse the static thresholds from Experiment I ($\tau_H = 1.3$ nats, $\tau_{\mathrm{MSP}} = 0.4$) to probe robustness.

\textbf{Results.} Table~\ref{tab:exp2_arch} reports overall (Ov.) and target-group (Tgt.; re-asked subset) statistics.

\begin{table*}[!t]
\caption{Generalization across model architectures on MMLU-Pro using static thresholds ($\tau_H = 1.3$ nats, $\tau_{\mathrm{MSP}} = 0.4$). This table summarizes the performance statistics for Qwen3 30B, Qwen3 235B, and DeepSeek-v3.2-Reasoner. It reports the mean Entropy (Ent.) and mean Max Probability (Prob.) for the Overall dataset (Ov.) in Round 1 (R1) and Round 2 (R2). Additionally, it presents the Accuracy (Acc.) for both the Overall dataset and the Target Reasked group (Tgt.) across both rounds. The Reask Rate denotes the proportion of samples selected for re-evaluation.}
\label{tab:exp2_arch}
\centering
\scriptsize
\setlength{\tabcolsep}{2pt}
\resizebox{\textwidth}{!}{%
\begin{tabular}{lrrrrrrrrr}
\hline\hline
Model & Ov. R1 Ent. & Ov. R1 Prob. & Ov. R2 Ent. & Ov. R2 Prob. & Ov. R1 Acc. & Ov. R2 Acc. & Tgt. R1 Acc. & Tgt. R2 Acc. & Reask Rate \\
\hline
Qwen3 30B & 1.2881 & 0.4292 & 1.3041 & 0.4300 & 76.37\% & 77.25\% & 73.41\% & 74.94\% & 57.50\% \\
Qwen3 235B & 1.4160 & 0.3753 & 1.4414 & 0.3752 & 81.44\% & 82.58\% & 77.78\% & 79.53\% & 65.24\% \\
DeepSeek V3.2 & 0.2456 & 0.7546 & 0.1917 & 0.7872 & 75.99\% & 78.70\% & 31.68\% & 44.61\% & 20.96\% \\
\hline\hline
\end{tabular}%
}
\end{table*}

On DeepSeek-v3.2-Reasoner, the re-ask rate is low (20.96\%), yet the target-group accuracy increases substantially (0.3168$\to$0.4461), suggesting that uncertainty-triggered re-asking can act as a compute-efficient ``safety net'' for reasoning-oriented inference. In contrast, the Qwen models trigger re-asking frequently (57.50\% and 65.24\%), but obtain only modest overall gains (0.7637$\to$0.7725 and 0.8144$\to$0.8258). This pattern is consistent with a reduced ``latent capability gap'' on a saturated benchmark regime, where high entropy may reflect intrinsic ambiguity rather than a correctable reasoning error.

\textbf{Implications.} These cross-model differences suggest that static thresholds are not universally compute-efficient. A natural next step is model-specific calibration or dynamic uncertainty scaling (e.g., entropy normalization against a model's baseline distribution), which can tune the accuracy--compute trade-off per architecture.

\subsection{Experiment III: Threshold Sweep and Parameter Choices}
\label{sec:threshold_sweep}

To understand how sensitive Reinforcement Inference is to the uncertainty thresholds, we ran a sweep over 16 threshold pairs on MMLU-Pro:
$$\tau_H\in\{0.7,0.9,1.1,1.3\}\ \text{nats},\qquad \tau_{\mathrm{MSP}}\in\{0.3,0.4,0.5,0.6\}.$$

\begin{figure}[!t]
\centering
\includegraphics[width=\linewidth]{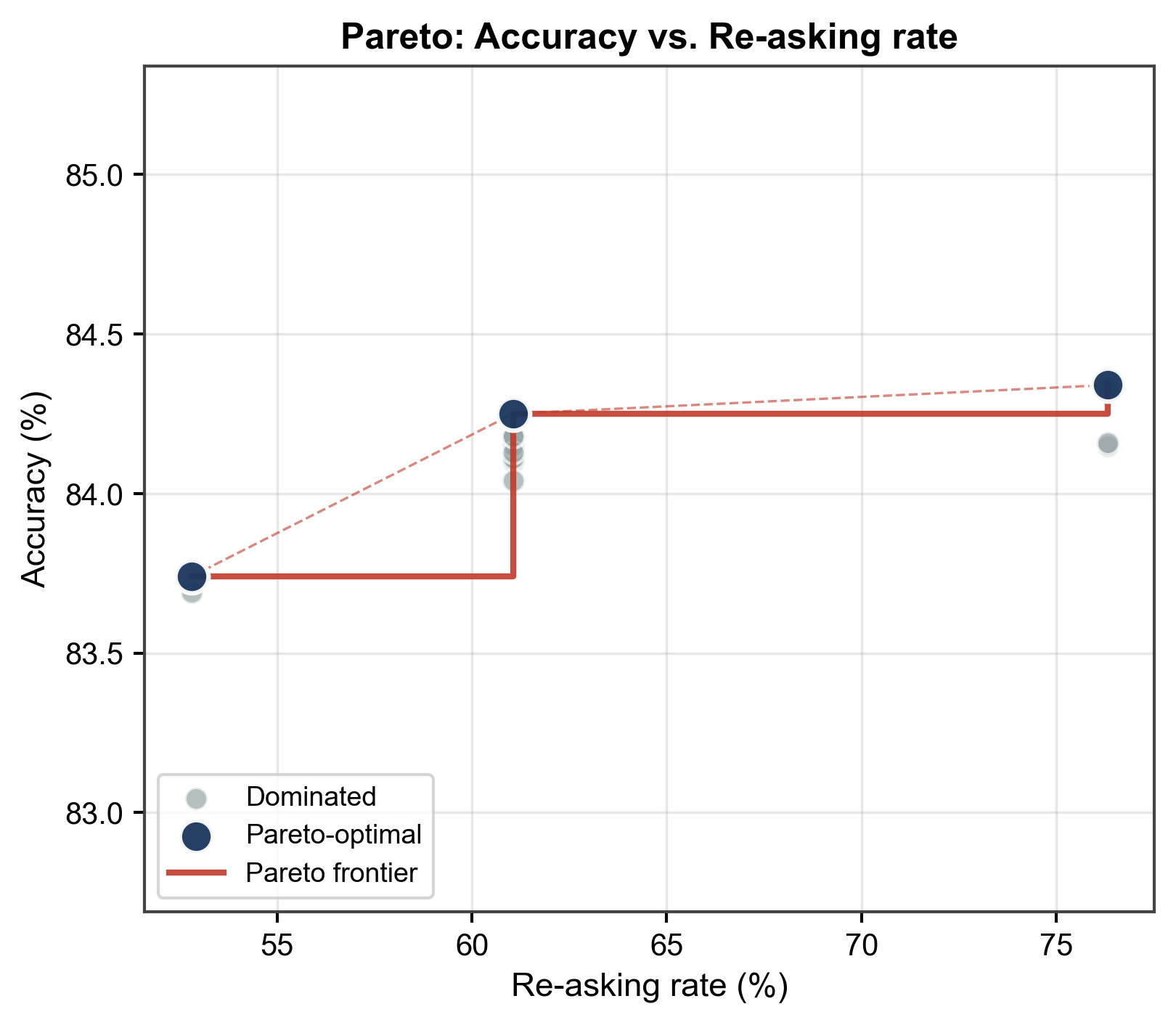}
\caption{Accuracy--compute trade-off from the 16-run sweep. Each point is one threshold pair $(\tau_H,\tau_{\mathrm{MSP}})$, plotted by its re-ask rate (x-axis) and final overall accuracy (y-axis).}
\label{fig:16exp_pareto}
\end{figure}

\textbf{How the Pareto curve is constructed.} Each threshold pair is a point $(r,a)$, where $r$ is the re-ask rate (compute) and $a$ is final accuracy. We define dominance as $r_1\le r_2$ and $a_1\ge a_2$ (strict in at least one), and the Pareto frontier in Figure~\ref{fig:16exp_pareto} is the set of non-dominated points. We compute it by sorting runs by increasing $r$ and retaining a point whenever it achieves a new best accuracy at that (or lower) compute.

The 16-point grid is guided by the first-round entropy/MSP distributions (Figure~\ref{fig:uncertainty}) to bracket where correct and incorrect cases overlap most, enabling a focused sweep over the compute--accuracy trade-off.

\textbf{Important caveat (threshold selection as an engineering aid).} The 16-run sweep is guided by the first-round outcome distributions (i.e., after evaluation we can observe which instances are correct/incorrect and how entropy/MSP behave). The model itself never sees any ground-truth labels; however, using labeled outcomes on the same benchmark to motivate an operating region can be viewed as a form of ``privileged'' information for \emph{policy selection}. We therefore treat this sweep primarily as an \emph{optimization/validation aid} to identify stable, compute-efficient operating points and to inform future cross-domain and cross-dataset inference policies. A stricter protocol would select \(\tau_H,\tau_{\mathrm{MSP}}\) on a held-out validation split (or cross-validation) and only then report test performance.

Table~\ref{tab:threshold_sweep} summarizes the 16 runs (accuracy and re-ask rate). Unless otherwise noted, we use $\tau_H=0.8$ nats and $\tau_{\mathrm{MSP}}=0.6$ as our main configuration; this setting yields 84.03\% final accuracy with 61.06\% re-asking on MMLU-Pro.

\begin{table}[!t]
\caption{Threshold sweep on MMLU-Pro (16 runs). Each run uses trigger $\mathbb{I}[H>\tau_H\lor\mathrm{MSP}<\tau_{\mathrm{MSP}}]$.}
\label{tab:threshold_sweep}
\centering
\footnotesize
\setlength{\tabcolsep}{4pt}
\begin{tabular}{lrrrr}
\hline\hline
Run & $\tau_H$ (nats) & $\tau_{\mathrm{MSP}}$ & Re-ask (\%) & Acc. (\%) \\
\hline
ent0.7\_mp0.3 & 0.7 & 0.3 & 61.06 & 84.18 \\
ent0.7\_mp0.4 & 0.7 & 0.4 & 61.06 & 84.10 \\
ent0.7\_mp0.5 & 0.7 & 0.5 & 61.06 & 84.04 \\
ent0.7\_mp0.6 & 0.7 & 0.6 & 76.31 & 84.15 \\
ent0.9\_mp0.3 & 0.9 & 0.3 & 61.06 & 84.14 \\
ent0.9\_mp0.4 & 0.9 & 0.4 & 61.06 & 84.11 \\
ent0.9\_mp0.5 & 0.9 & 0.5 & 61.06 & 84.15 \\
ent0.9\_mp0.6 & 0.9 & 0.6 & 76.31 & 84.33 \\
ent1.1\_mp0.3 & 1.1 & 0.3 & 52.80 & 83.74 \\
ent1.1\_mp0.4 & 1.1 & 0.4 & 61.06 & 84.18 \\
ent1.1\_mp0.5 & 1.1 & 0.5 & 61.06 & 84.13 \\
ent1.1\_mp0.6 & 1.1 & 0.6 & 76.31 & 84.16 \\
ent1.3\_mp0.3 & 1.3 & 0.3 & 52.80 & 83.69 \\
ent1.3\_mp0.4 & 1.3 & 0.4 & 61.06 & 84.25 \\
ent1.3\_mp0.5 & 1.3 & 0.5 & 61.06 & 84.18 \\
ent1.3\_mp0.6 & 1.3 & 0.6 & 76.31 & 84.34 \\
\hline\hline
\end{tabular}
\end{table}

\subsection{Experiment IV: Cross-Dataset Transfer: Zero-shot MMLU (4-way)}
\label{sec:mmlu_transfer}

To test generalization beyond MMLU-Pro (10-way), we transferred the normalized entropy threshold implied by \(\tau_H=1.3\) nats (10-way) to standard 4-way MMLU. Using
$$\bar{H}_\tau = \frac{1.3}{\ln 10}\approx 0.565\quad\Rightarrow\quad \tau_H^{(K=4)}=\bar{H}_\tau\ln 4\approx 0.78\ \text{nats},$$
with \(\tau_{\mathrm{MSP}}=0.4\), we evaluated 14,042 zero-shot multiple-choice questions.

Overall, accuracy changes only marginally on standard MMLU: 91.134\% $\to$ 91.184\% (+0.050pp). Across 57 subjects, 16 improve and 11 degrade, with the rest unchanged. Thus, while uncertainty-triggered re-asking yields large gains on harder MMLU-Pro, its transfer to an already strong regime ($\sim$91\%) provides only minor net benefit and may cause small subject-level regressions.

\begin{table}[!t]
\caption{Zero-shot MMLU transfer (4-way): normalized-entropy mapping from 10-way MMLU-Pro and MSP threshold kept fixed.}
\label{tab:mmlu_transfer}
\centering
\footnotesize
\setlength{\tabcolsep}{4pt}
\begin{tabular}{lr}
\hline\hline
Metric & Value \\
\hline
Dataset size & 14,042 questions \\
$K$ (choices) & 4 \\
Entropy threshold $\tau_H$ (nats) & 0.78 \\
MSP threshold $\tau_{\mathrm{MSP}}$ & 0.4 \\
Baseline accuracy & 91.134\% \\
TR (Reinforcement Inference) accuracy & 91.184\% \\
Improved / Degraded subjects & 16 / 11 (out of 57) \\
\hline\hline
\end{tabular}
\end{table}

\section{Discussion}

Our findings demonstrate uncertainty-aware inference control at scale. Reinforcement Inference enables DeepSeek-v3.2 to behave more like a careful human: when uncertain, pause and rethink. It leverages internal uncertainty signals \cite{hendrycks2017baseline, guo2017calibration} to improve performance without retraining or external feedback. Below we summarize why the second pass helps and when it can fail, motivating the limitations and future directions that follow.

\begin{itemize}
    \item \textbf{Latent Knowledge Utilization:} The correct option often has non-negligible probability even when the first-pass answer is wrong, suggesting latent knowledge is present but underused. The reinforcement cue (“you weren’t confident, so think harder”) often elicits deeper reasoning and recovers the correct answer.
    \item \textbf{Change of Decoding Dynamics:} When the model is nearly indifferent among a few options, the second-pass cue encourages considering alternatives and justifying choices, often breaking ties in favor of the correct option. This echoes why CoT can help \cite{kojima2022large, yao2023tree}: explicit reasoning can resolve internal ambiguity. RI induces this extra deliberation only on uncertain cases.
    \item \textbf{Model’s Self-Monitoring:} Re-asking is not risk-free: 107 initially correct answers flipped to wrong (1.46\% of re-asked questions), though the net effect remained strongly positive (2,912 wrong\(\to\)correct vs.\ 107 correct\(\to\)wrong). A key limitation is that extremely overconfident mistakes may not trigger re-asking.
    \item \textbf{Generality:} Although we focus on multiple-choice QA, the idea can extend to open-ended tasks by triggering re-asking when uncertainty is high (e.g., low likelihood of the top completion or high entropy over continuations). This is harder without discrete options, but prior work defines entropy in open generation by clustering semantically equivalent answers \cite{kuhn2023semantic}. A practical variant is: answer once, estimate confidence (e.g., log-likelihood or sampling variance), then request justification and revision when confidence is low.
    \item \textbf{Efficiency Considerations:} Our method adds at most one extra call on flagged queries. In targeted re-asking (TR), this is 7,347 extra calls for 12,032 queries (61\% more total calls) while achieving 99.6\% of the gain from uniform re-asking (UR). This is far cheaper than self-consistency (5--10 calls per query; 500\%--1000\% overhead) \cite{wang2023self, lewkowycz2022solving} or always-on CoT. In interactive settings, latency increases mainly on hard, uncertain queries, and thresholds can be tuned for speed--accuracy trade-offs.
    \item \textbf{Comparison to “Step-Back” Prompting:} ``Step-back'' prompts ask models to double-check, but applying them universally can waste compute and even hurt performance. Our method formalizes this idea by tying it to an uncertainty metric: we trigger reflection only when the model’s probabilities indicate hesitation. This selective trigger is central to the efficiency and avoids unnecessary second-guessing on confident (often correct) cases.
\end{itemize}

\section{Limitations}

Despite the encouraging results, our study has clear limitations that suggest next steps: improving uncertainty calibration across model families, extending entropy-triggered deliberation beyond multiple-choice tasks, and making the second-pass prompting policy more robust while remaining compute-aware. We discuss these limitations below.

\begin{itemize}
    \item \textbf{Dependency on Model Calibration:} Effectiveness is not uniform across architectures and may depend on multiple factors, including training provenance (e.g., possible benchmark exposure), model architecture, and baseline performance ceilings. For example, the minimal gains on Qwen3 despite high re-asking rates could reflect prior exposure to MMLU-Pro during post-training/alignment, but could also indicate that the model is already strong (or more decisive) on this benchmark, leaving limited marginal room for improvement. In contrast, DeepSeek-v3.2 and v3.2-Reasoner show substantial improvements. Overall, Reinforcement Inference is most useful when the model exhibits genuine uncertainty and there is headroom for correction; deployments should therefore consider both training history and baseline accuracy when selecting benchmarks and operating points.
    \item \textbf{Multiple-Choice Focus:} Our implementation targets multiple-choice QA, where correctness is clear and uncertainty is naturally defined over a discrete option set. Extending to open-ended tasks (free-form QA, code generation) requires more careful uncertainty definitions and alternative-solution generation (e.g., sampling/ensembles to estimate entropy), which we do not study here.
    \item \textbf{Prompt Specificity:} We adopt a single fixed second-pass prompt and do not systematically explore alternative phrasings. While different wordings may yield further improvements, they could also increase verbosity or introduce additional tuning effects. Moreover, incorporating few-shot corrective examples may enhance performance, but would depart from the zero-shot setting we aim to preserve for comparability and simplicity.
    \item \textbf{Anchoring and prompt ablations:} The second-pass prompt includes the first-pass output (``Your previous answer was: ...'') and an explicit entropy cue, which may introduce anchoring (defending or over-correcting relative to the prior answer). While our prompt-only ablation (Section~\ref{sec:prompt_only_ablation}) shows generic CoT encouragement alone does not explain gains, future work should further ablate removing the prior output and/or entropy cue to isolate each component.
    \item \textbf{No Adversarial or Trick Questions:} We do not test adversarial cases designed to exploit overconfidence or mislead uncertainty signals. In such settings, the method may not trigger (confidently wrong) or may still fail if reasoning is flawed. Addressing overconfident mistakes likely requires external verification or improved self-uncertainty training; identifying such cases remains open.
\end{itemize}

\section{Future Work}

Encouraging results from this initial study motivate several directions for future work. The directions below emphasize generality, robustness, and deployable system policies:

\begin{itemize}
    \item \textbf{Broader Evaluation:} We plan to evaluate Reinforcement Inference across more tasks and models. Testing other advanced models and newer open-source systems would clarify whether uncertainty-triggered reasoning generalizes. When raw probabilities are unavailable, one may need alternatives such as prompting confidence estimates, though these can be unreliable without calibration. Expanding to more benchmarks (e.g. other MMLU categories, reasoning-heavy datasets like GSM8K, or coding challenges) will further test generality.
    \item \textbf{Dynamic Thresholds and Policies:} In deployment, thresholds (e.g., \(\tau\)) need not be fixed: a system could raise \(\tau\) when second passes yield little benefit or lower it when maximizing accuracy is prioritized. More generally, one could learn a lightweight meta-policy that predicts whether a second pass will help from features of the probability distribution and question.
    \item \textbf{Multi-Step Reinforcement and Verification:} We currently allow only one extra pass. If the model remains uncertain after the second attempt, one could allow a small number of additional iterations while enforcing a strict cap to avoid loops. In critical applications, this can be paired with verification or abstention: (i) produce an initial answer, (ii) re-ask if uncertain, and (iii) if uncertainty remains high after reinforcement, refuse or escalate to a human. Measuring how often uncertainty persists after the second pass---and whether additional iterations or verification helps---is an open question.
    \item \textbf{Applications in Generative Tasks:} Although we focused on Q\&A, uncertainty-triggered self-correction may apply to generative settings such as code synthesis or long-form writing. High-entropy regions during generation could trigger a ``double-check'' or alternative-solution attempt to improve correctness or coherence. Understanding how to detect and act on uncertainty in open-ended generation is an intriguing direction.

    \item \textbf{Extending Reinforcement Inference to VLA and Robotics:} A natural extension is to move from discrete answer selection to vision--language--action (VLA) and robotic control models \cite{brohan2023rt2}, which are increasingly decoder-based and autoregressive and thus admit uncertainty-triggered refinement over action sequences. Entropy over the action distribution can serve as an execution-uncertainty signal: high entropy corresponds to diffuse, unstable action proposals, while low entropy indicates convergent trajectories. Reinforcement Inference could then allocate extra refinement only when this uncertainty is high, encouraging trajectories to concentrate toward high-certainty regions---a pattern that parallels human motor control, where early motion is exploratory and later motion becomes precise. Compared to imitation learning \cite{ross2011dagger} or reward-driven reinforcement learning, this entropy-constrained autoregressive perspective directly targets reducing execution uncertainty and remains compatible with modern visuomotor policy learning methods \cite{chi2023diffusion}.

    \item \textbf{Entropy-Guided Training:} Our results suggest models often \emph{know} when they are unsure, but are not trained to use that signal by default. A natural direction is to integrate uncertainty into training, e.g., adding an auxiliary objective that encourages low entropy (high confidence) only when correct and higher entropy when genuinely uncertain, improving calibration. Recent work emphasizes that LLMs must learn what they do not know \cite{kapoor2024large}. Rather than relying solely on post-hoc calibration fine-tuning \cite{ouyang2022training, guo2017calibration}, one could train with penalties for incorrect high-confidence predictions from the outset, yielding models that are both more reliable and better suited for reinforcement inference due to more trustworthy uncertainty signals.
\end{itemize}

\section{Conclusion}

We introduced Reinforcement Inference, an inference-time method that leverages a model’s uncertainty to selectively trigger a second, more deliberate attempt.

On 12,032 MMLU-Pro multiple-choice questions across 14 subjects, DeepSeek-v3.2 improves from 60.72\% to 84.03\% (+23.31\%) while re-asking 61.06\% of questions; re-asking 100\% yields only +0.10\% additional accuracy over targeted re-asking (TR), indicating that uncertainty-based triggering captures most of the gain with substantially less compute.

More broadly, our framework is naturally aligned with the autoregressive nature of modern decoder-based models: because outputs are generated token by token from a learned conditional distribution, entropy and related confidence measures arise directly during generation and can be used as \emph{control signals}. Reinforcement Inference exploits this coupling by turning distributional uncertainty into an adaptive inference policy---allocating extra compute only when the model hesitates---to improve reliability without changing model weights. This perspective complements broader efforts to align and control LLM behavior and motivates further work on uncertainty-aware inference and training.

\end{document}